\documentclass[runningheads]{llncs}

\usepackage{tikz}
\usepackage{xcolor}
\usepackage{graphicx}
\begin{document}

\title{Event-Based Vision in Space: Applications, Trends, and Future Directions}

\author{
Luigi Capogrosso\inst{1} \and
Pietro Bonazzi\inst{2} \and
Michele Magno\inst{2,1}
}

\authorrunning{L. Capogrosso et al.}

\institute{
Interdisciplinary Transformation University of Austria, \email{name.surname@it-u.at}
\and
ETH Zurich, Zurich, \email{name.surname@pbl.ee.ethz.ch}}

\AddToHookNext{shipout/foreground}{%
\begin{tikzpicture}[overlay, remember picture]
    \node at ([yshift=-1.3cm]current page.north) {
        \normalsize\textcolor{gray}{This paper has been accepted for publication at the}
    };
    \node at ([yshift=-1.8cm]current page.north) {
        \normalsize\textcolor{gray}{XXIV Annual Conference on Sensors and Microsystems (AISEM), Catania, Italy, 2026}
    };
\end{tikzpicture}%
}

\maketitle

\begin{abstract}
Earth Observation (EO) is undergoing a significant transformation driven by the deployment of novel sensing technologies.
Traditional frame-based optical sensors often struggle with motion blur, high power consumption, and extreme data redundancy in challenging orbital environments.
In contrast, event-based sensors, also known as neuromorphic cameras, offer a bio-inspired asynchronous approach.
By capturing only local illumination changes, they provide microsecond temporal resolution, an extremely high dynamic range, and exceptional energy efficiency.
Although the use of these sensors is rapidly expanding from terrestrial systems to orbital platforms, the scientific literature surrounding their space-based applications remains heavily fragmented.
To bridge this gap, this article presents a comprehensive review of the state-of-the-art in event-based vision in the space domain.
Based on the retrieved literature, we introduce a taxonomy structured around four primary domains: 1) atmospheric and high-speed observation; 2) environmental monitoring and change detection; 3) operational support and onboard processing; and 4) geospatial modeling and predictive analysis.
As a result, this survey highlights that neuromorphic engineering is far more than a supplementary imaging technique; it is a paradigm shift that can be used to directly address critical bottlenecks in modern remote sensing and sustainable space exploration.
\end{abstract}

\section{Introduction} \label{sec:intro}

In the last ten years, the field of Earth Observation (EO) has undergone a major transformation with the introduction of novel sensing technologies \cite{Toth2016,Tuia2025,Duggan2025}.
Traditional optical sensors capture images at fixed frame rates, an approach that often results in motion blur, high power consumption, and excessive generation of redundant data \cite{Gallego2022}.
Event-based sensors, also known as neuromorphic cameras, offer a fundamentally different approach \cite{Chakravarthi2024}.
Inspired by the human retina, these sensors respond only to local changes in brightness \cite{Yang2024}.
They generate data asynchronously, producing a continuous stream of events with microsecond temporal resolution, high dynamic range, and significantly lower power requirements \cite{Lichtsteiner2008,Capogrosso2026b}.
As illustrated on the left side of Figure \ref{fig:teaser-and-taxonomy.pdf}, adapted from \cite{Chakravarthi2024}, this asynchronous paradigm completely eliminates redundant background data typical of standard frame-based vision.

\emph{\textbf{Motivations for this paper.}}
These unique characteristics make event-based sensors particularly suitable for the challenging environments of space and high-speed aerial observation \cite{McHarg2024}.
Although event-based vision has established a solid foundation in terrestrial robotics \cite{Gallego2022,Chakravarthi2024}, its emerging integration into space platforms is currently accompanied by highly fragmented academic literature.
Currently, there is a lack of a comprehensive review that consolidates how event-based and neuromorphic hardware and algorithms are being applied specifically in space exploration.

\emph{\textbf{Scientific contribution.}}
To address this gap, this article synthesizes the state-of-the-art literature into a unified taxonomy, categorizing the emerging applications of neuromorphic sensing in the space domain.
As a result, the main contribution is to define a clear taxonomy of how this technology is currently being deployed and researched, highlighting its potential to solve long-standing challenges in remote sensing and satellite operations.
Furthermore, this review acts as a bridge between the neuromorphic engineering community and remote sensing professionals, establishing a foundational reference for future interdisciplinary development.

\begin{figure}[t!]
    \centering
    \includegraphics[width=\linewidth]{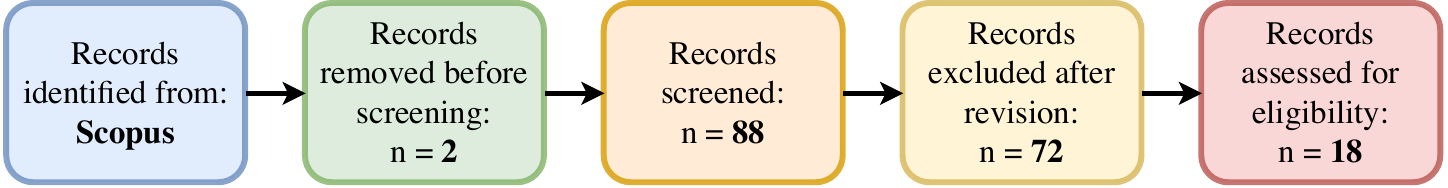}
    \caption{Flowchart of the literature search and selection process adapted from the PRISMA guidelines \cite{Page2021}.}
    \label{fig:prisma-light.pdf}
\end{figure}
To ensure a comprehensive overview of the relevant literature, an accurate search was conducted using a ``PRISMA-light'' procedure \cite{Page2021}, detailed in Figure \ref{fig:prisma-light.pdf}.
We use the term ``PRISMA-light'' to indicate the intentional omission of certain steps because our scope is limited to event-based vision within the space domain.
Since comprehensive surveys already exist for the wider field of event-based vision \cite{Gallego2022,Chakravarthi2024}, a complete systematic review of the entire literature is unnecessary to define a targeted taxonomy.

Specifically, the article selection strategy was carefully constructed to retrieve documents that include the term \texttt{`Earth Observation'} in conjunction with the key concepts related to this new sensing paradigm, \emph{i.e.}, \texttt{`event-based'}, \texttt{`neuromorphic'}, or \texttt{`spiking neural networks'}.
These terms were searched in the titles, abstracts, and keywords of the Scopus-indexed records.
To ensure consistency and accessibility, the articles retrieved were limited to publications in English, and all studies had to be published in peer-reviewed journals or conference proceedings.
This initial query identified 90 records.
As shown in Figure \ref{fig:prisma-light.pdf}, 2 records were removed before the screening phase, leaving 88 records to be screened.

The search yielded numerous false positives because, for example, the term \texttt{`event'} is frequently used in traditional remote sensing to describe standard natural occurrences such as \texttt{`rain events'} or \texttt{`disaster events'}.
Thus, we filtered the retrieved articles according to their content.
During this review, we excluded traditional studies, focusing only on research that explicitly uses neuromorphic hardware, event-based vision sensors, or spiking neural network architectures.
Following this stage, 72 records were excluded, resulting in a final set of 18 records that were fully assessed for eligibility and included to form the core of our taxonomy.

The analysis of the filtered literature yields a four-pillar taxonomy.
In particular, the remaining articles naturally clustered into the following four categories:
\begin{enumerate}
    \item Studies exploiting the extreme temporal resolution of event cameras naturally formed the \emph{atmospheric and high-speed observation} category.
    \item Research utilizing the sensors' high dynamic range for robust surface analysis is grouped into \emph{environmental monitoring and change detection}.
    \item The literature on low-latency and energy-efficient edge computing defined the category of \emph{operational support and onboard processing}.
    \item Finally, research on event-driven asynchronous algorithms formed the \emph{geospatial modeling and predictive analysis} category.
\end{enumerate}
By organizing the literature into these four distinct categories, this review establishes a structured and comprehensive overview of the emerging field of event-based vision for space applications.
\section{Background} \label{sec:background}

Event cameras, also known as neuromorphic or Dynamic Vision Sensors (DVS), are bio-inspired vision sensors \cite{AliAkbarpour2024}.
Unlike conventional frame-based cameras that capture intensity images at a fixed frame rate, each pixel of an event camera independently and asynchronously measures changes in logarithmic intensity and generates an event only when the change exceeds a predefined contrast threshold \cite{Gallego2022}.

Formally, an event $e_k = (x_k, y_k, t_k, p_k)$ is triggered at pixel $(x_k, y_k)$ and timestamp $t_k$ with polarity $p_k \in \{-1, +1\}$ whenever the log-intensity $L(x,y,t) = \log I(x,y,t)$ changes by at least a contrast threshold $C$:
\begin{equation}
L(x_k,y_k,t_k) - L(x_k,y_k,t_{k-1}) = p_k C\;,
\end{equation}
where $t_{k-1}$ is the time of the previous event at the same pixel.
This asynchronous, sparse, event-driven readout yields microsecond-level temporal resolution (typically $1\,\mu$s), a dynamic range exceeding 120--140~dB, negligible motion blur, and significantly lower power and bandwidth consumption compared to traditional cameras \cite{Lichtsteiner2008,Gallego2022}.

The first silicon implementation of such a sensor was introduced in \cite{Lichtsteiner2008}, followed by the improved DVS128 and subsequent commercial devices, \emph{e.g.}, iniVation DAVIS, Prophesee Metavision, and Sony IMX636.
Early event cameras suffered from fixed-pattern noise and limited spatial resolution, but modern sensors have reached VGA resolution with improved noise characteristics and and integrated Active Pixel Sensor (APS) circuitry that enables the output of grayscale frames (as seen in DAVIS-type cameras).

Due to their unique properties, event cameras have enabled breakthroughs in high-speed vision tasks such as motion estimation \cite{Rebecq2021,Shiba2022,Zhao2025}, tracking \cite{Messikommer2023,Bonazzi2024,Bonazzi2025}, SLAM \cite{Rebecq2018,Vidal2018}, and 3D reconstruction \cite{Zhou2018,Li2025} under challenging lighting conditions and fast motion. 
Comprehensive surveys of the field and its algorithms are provided by \cite{Gallego2022} and more recent overviews of hardware innovations are available in \cite{Chakravarthi2024}. 
\section{Taxonomy} \label{sec:methodology}

\begin{figure}[t!]
    \centering
    \includegraphics[width=\linewidth]{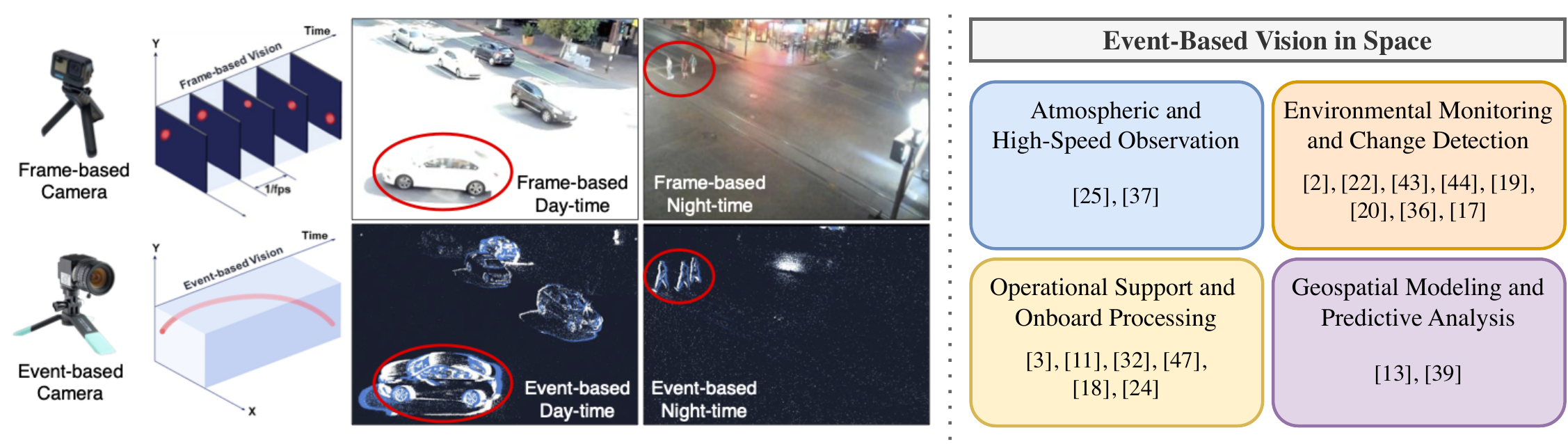}
    \caption{The left panel illustrates a conceptual comparison from \cite{Chakravarthi2024} between traditional frame-based cameras and event-based sensors.
    On the right, our proposed taxonomy outlining the current applications of event-based and neuromorphic sensing in space-based applications, categorized into the distinct research domains with their corresponding core literature.}
    \label{fig:teaser-and-taxonomy.pdf}
\end{figure}

The applications of event-based sensors and neuromorphic computing in space-domain applications can be categorized into the distinct areas summarized on the right side of Figure \ref{fig:teaser-and-taxonomy.pdf}.

\subsection{Atmospheric and High-Speed Observation}
The first category involves the observation of atmospheric and high-speed phenomena.
In this domain, the extreme speed and precise timing of event cameras are used to capture rapid natural events.
A particular real-world example is the Falcon Neuro mission, which uses event-based sensors in space to observe the precise timing and distribution of lightning strikes \cite{McHarg2024}.
Similarly, researchers are currently developing neuromorphic event-based imaging spectroscopy to observe vehicles flying at hypersonic speeds, fully exploiting the fast reaction times of the sensor \cite{Sopek2026}.

\subsection{Environmental Monitoring and Change Detection}
The second category covers environmental monitoring, land use analysis, and change detection.
Given that event cameras possess a high dynamic range, they are ideally suited for reliably detecting features on the Earth's surface under extreme lighting conditions.
To obtain the clearest possible images from these raw sensors, researchers have developed methods such as density-invariant contrast maximization, which significantly improves the visual quality of neuromorphic EO \cite{Arja2023}.
Building on this, recent frameworks like ERS-HDRI leverage the synergy between event streams and conventional frames to reconstruct high-quality high-dynamic-range (HDR) remote sensing imagery, effectively overcoming the limitations of poorly exposed orbital captures \cite{Li2024}.
To further enhance observation capabilities, recent aerospace missions, such as the VANTAGE experiment, are pioneering sensor fusion techniques that overlap the field of view of traditional visual images with neuromorphic event-based sensors to intelligently correlate multi-spectral data with high-speed intensity changes \cite{Wildenstein2025}.

Spiking neural networks \cite{Eshraghian2023}, \emph{i.e.}, models that process information via asynchronous bio-inspired spike trains, are increasingly paired with these sensors.
In this regard, attention-based spiking neural networks have recently achieved state-of-the-art results in the super-resolution of remote sensing imagery, actively mitigating the information loss typical of degraded satellite textures while maintaining high computational efficiency \cite{Xiao2025}.
For example, they have been successfully tested for the classification of energy-efficient land cover and land use directly on board satellites \cite{Kucik2021}, as well as for general onboard satellite image classification tasks \cite{Lemaire2020}.
To detect sudden environmental changes, recent studies have introduced a physics-aware training-free model based on neuromorphic networks to track rapid changes on the surface of the Earth \cite{Smith2025}.
Additionally, these low-power processors enable rapid cloud-cover detection for small satellites, optimizing data acquisition by filtering obscured scenes prior to high-resolution imaging \cite{Kadway2023}.

\subsection{Operational Support and Onboard Processing}
A third significant application area relates to the operational support of the spacecraft and data processing directly on the satellite.
In particular, event cameras show great promise for star tracking, which helps satellites figure out exactly where they are pointing even when the spacecraft is shaking or vibrating, as demonstrated by the e-STURT dataset \cite{Bagchi2026}.
In parallel with star tracking, event cameras are now being actively deployed for sun sensing, as demonstrated by the introduction of the Sun-E benchmark \cite{Dolan2026}, which allows spacecraft to leverage the sensors' extreme dynamic range for precise attitude determination even under direct solar illumination.
This neuromorphic approach also provides a robust framework for autonomous relative localization in future space exploration missions \cite{Salah2024}.

Beyond navigation, spiking neural networks can also help manage complex daily tasks on a satellite.
For example, they have been applied to efficiently schedule multi-point imaging missions for EO satellites \cite{Yao2025}.
Furthermore, they are used to compress images without losing quality, reducing the power consumption of small satellites before data are transmitted back to Earth \cite{Kahali2023}.
The practical feasibility of these onboard tasks has recently been validated through rigorous energy efficiency analyzes of commercial neuromorphic edge processors, demonstrating their readiness to handle complete EO workflows directly in orbit without the need to transmit unprocessed data to ground stations \cite{Lunghi2025}.

\subsection{Geospatial Modeling and Predictive Analysis}
Finally, the taxonomy presents advanced modeling of hydrological and environmental data.
For example, spiking neural networks combined with remote sensing data have been shown to be intelligent and highly effective in predicting how soil moisture varies over time in different regions \cite{ElMaachi2025}.
Furthermore, event-based serverless systems in the cloud are being adopted to run computationally intensive environmental models, such as river system analysis, in a highly efficient and accessible way \cite{Tom2026}.

Through this classification, it becomes evident that event-based sensing is not merely a replacement for standard optical cameras, but a completely new operational approach for both observing the Earth and navigating the platforms that orbit it.
\section{Discussion \& Concluding Remarks} \label{sec:conclusions}

The integration of event-based sensor platforms represents a significant paradigm shift in how we monitor our planet and navigate spacecraft.
As demonstrated through the proposed taxonomy, neuromorphic engineering offers promising solutions to some of the most persistent bottlenecks in traditional remote sensing.
By capturing only dynamic changes in a scene, event cameras drastically reduce redundant data generation, lower power consumption, and provide the microsecond reaction times necessary to capture high-speed atmospheric phenomena \cite{Gallego2022,Nishiguchi2024}.
Furthermore, when coupled with spiking neural networks, these sensors enable highly efficient onboard data processing \cite{Roy2019,Yang2026}.
This allows satellites to make autonomous decisions on the edge rather than transmitting massive unprocessed datasets back to Earth, a capability driven by advanced hardware and software co-design strategies \cite{Capogrosso2024,Capogrosso2026a}.

Despite these significant advantages, several critical challenges must be addressed before neuromorphic technology can reach widespread adoption in space.
The foremost limitation is the maturity of the hardware \cite{Lunghi2025}.
Although early deployments, such as the Falcon Neuro mission, have demonstrated the viability of event sensors in low Earth orbit, there remains a pressing need to develop robust, radiation-hardened neuromorphic cameras and processors capable of withstanding long-term deep-space missions \cite{Esa2024}.
In addition, there is a software gap.
The asynchronous and spatially sparse nature of event data precludes the direct application of traditional computer vision algorithms designed for standard optical frames \cite{Cuadrado2023}.
Researchers must continue to develop native, event-driven algorithms and training methodologies for spiking neural networks, which currently lag behind standard deep learning models in terms of ease of use and standardization, as demonstrated in \cite{Shi2024}.

Looking ahead, interdisciplinary collaboration between aerospace engineers and neuromorphic researchers will be essential.
A primary focus for future research should be the creation of large-scale and standardized event-based Earth observation datasets, such as the Sun-E introduced in \cite{Dolan2026}.
Just as traditional optical datasets accelerated the development of standard artificial intelligence, open-source neuromorphic datasets will be crucial to benchmark new models for tasks such as cloud detection, land cover classification, and spacecraft tracking.
Another crucial research direction is sensor fusion to identify optimal ways to combine high-speed temporal data from event cameras with rich multispectral spatial data from traditional remote sensing instruments \cite{Gehrig2019,Shi2025}.

In conclusion, event-based sensing is increasingly becoming a highly practical tool for space exploration and environmental monitoring.
The literature reviewed in this article illustrates a rapidly evolving field that is actively solving real-world problems in satellite operations and geospatial modeling.
As launch costs decrease and small satellite constellations multiply, the demand for low-power, high-efficiency orbital sensors will only grow.
Neuromorphic engineering is uniquely positioned to meet this demand, providing a sustainable, intelligent, and real-time operational approach for the next generation of EO missions.

\section*{Acknowledgments}
The research was funded by the Swiss National Science Foundation (Grant 219943).

\bibliographystyle{plain}
\bibliography{bibliography}

@Article{Toth2016,
  author    = {Toth, Charles and Jóźków, Grzegorz},
  journal   = {ISPRS Journal of Photogrammetry and Remote Sensing},
  title     = {{Remote sensing platforms and sensors: A survey}},
  year      = {2016},
  pages     = {22--36},
  volume    = {115},
}

@Article{Tuia2025,
  author    = {Tuia, Devis and Schindler, Konrad and Demir, Begüm and Zhu, Xiao Xiang and Kochupillai, Mrinalini and Džeroski, Sašo and van Rijn, Jan N. and Hoos, Holger H. and Frate, Fabio Del and Datcu, Mihai and Markl, Volker and Saux, Bertrand Le and Schneider, Rochelle and Camps-Valls, Gustau},
  journal   = {IEEE Geoscience and Remote Sensing Magazine},
  title     = {{Artificial Intelligence to Advance Earth Observation: A review of models, recent trends, and pathways forward}},
  year      = {2025},
  number    = {4},
  pages     = {119--141},
  volume    = {13},
}

@Article{Duggan2025,
  author    = {Duggan, Aidan and Andrade, Bruno and Afli, Haithem},
  journal   = {European Journal of Remote Sensing},
  title     = {{Advancing Earth observation: a survey on AI-powered image processing in satellites}},
  year      = {2025},
  number    = {1},
  volume    = {58},
}

@Article{Gallego2022,
  author    = {Gallego, Guillermo and Delbruck, Tobi and Orchard, Garrick and Bartolozzi, Chiara and Taba, Brian and Censi, Andrea and Leutenegger, Stefan and Davison, Andrew J. and Conradt, Jorg and Daniilidis, Kostas and Scaramuzza, Davide},
  journal   = {IEEE Transactions on Pattern Analysis and Machine Intelligence},
  title     = {{Event-Based Vision: A Survey}},
  year      = {2022},
  number    = {1},
  pages     = {154--180},
  volume    = {44},
}

@Article{Yang2024,
  author    = {Yang, Xu and Yao, Chunhe and Kang, Lei and Luo, Qian and Qi, Nan and Dou, Runjiang and Yu, Shuangming and Feng, Peng and Wei, Zhongming and Liu, Jian and Wang, Kaiyou and Wu, Nanjian and Liu, Liyuan},
  journal   = {IEEE Journal of Solid-State Circuits},
  title     = {{A Bio-Inspired Spiking Vision Chip Based on SPAD Imaging and Direct Spike Computing for Versatile Edge Vision}},
  year      = {2024},
  number    = {6},
  pages     = {1883--1898},
  volume    = {59},
}

@Article{Lichtsteiner2008,
  author    = {Lichtsteiner, Patrick and Posch, Christoph and Delbruck, Tobi},
  journal   = {IEEE Journal of Solid-State Circuits},
  title     = {{A 128$\times$128 120 dB 15 $\mu$s Latency Asynchronous Temporal Contrast Vision Sensor}},
  year      = {2008},
  number    = {2},
  pages     = {566--576},
  volume    = {43},
}

@Article{Capogrosso2026b,
  title   = {{Performance Analysis of Edge and In-Sensor AI Processors: A Comparative Review}},
  author  = {Capogrosso, Luigi and Bonazzi, Pietro and Magno, Michele},
  journal = {arXiv preprint arXiv:2603.08725},
  year    = {2026}
}

@InProceedings{Chakravarthi2024,
  author    = {Chakravarthi, Bharatesh and Verma, Aayush Atul and Daniilidis, Kostas and Fermuller, Cornelia and Yang, Yezhou},
  title     = {{Recent Event Camera Innovations: A Survey}},
  year      = {2024},
  booktitle = {European Conference on Computer Vision Workshops (ECCVW)},
}

@Article{McHarg2024,
  author    = {McHarg, Matthew G. and Jones, Imogen R. and Wilcox, Zachary and Balthazor, Richard L. and Marcireau, Alexandre and Cohen, Gregory},
  journal   = {Frontiers in Remote Sensing},
  title     = {{Falcon Neuro space-based observations of lightning using event-based sensors}},
  year      = {2024},
  volume    = {5},
}

@Article{Page2021,
  author    = {Page, Matthew J. and McKenzie, Joanne E. and Bossuyt, Patrick M. and Boutron, Isabelle and Hoffmann, Tammy C. and Mulrow, Cynthia D. and Shamseer, Larissa and Tetzlaff, Jennifer M. and Akl, Elie A. and Brennan, Sue E. and Chou, Roger and Glanville, Julie and Grimshaw, Jeremy M. and Hróbjartsson, Asbjørn and Lalu, Manoj M. and Li, Tianjing and Loder, Elizabeth W. and Mayo-Wilson, Evan and McDonald, Steve and McGuinness, Luke A. and Stewart, Lesley A. and Thomas, James and Tricco, Andrea C. and Welch, Vivian A. and Whiting, Penny and Moher, David},
  journal   = {Systematic Reviews},
  title     = {{The PRISMA 2020 statement: an updated guideline for reporting systematic reviews}},
  year      = {2021},
  number    = {1},
  volume    = {10},
}

@Article{AliAkbarpour2024,
  author    = {AliAkbarpour, Hadi and Moori, Ahmad and Khorramdel, Javad and Blasch, Erik and Tahri, Omar},
  journal   = {IEEE Sensors Reviews},
  title     = {{Emerging Trends and Applications of Neuromorphic Dynamic Vision Sensors: A Survey}},
  year      = {2024},
  pages     = {14--63},
  volume    = {1},
}

@Article{Rebecq2021,
  author    = {Rebecq, Henri and Ranftl, Rene and Koltun, Vladlen and Scaramuzza, Davide},
  journal   = {IEEE Transactions on Pattern Analysis and Machine Intelligence},
  title     = {{High Speed and High Dynamic Range Video with an Event Camera}},
  year      = {2021},
  number    = {6},
  pages     = {1964--1980},
  volume    = {43},
}

@InProceedings{Shiba2022,
  title     = {{Secrets of Event-Based Optical Flow}},
  author    = {Shiba, Shintaro and Aoki, Yoshimitsu and Gallego, Guillermo},
  booktitle = {European Conference on Computer Vision (ECCV)},
  year      = {2022}
}

@InProceedings{Zhao2025,
  title     = {Full-DoF Egomotion Estimation for Event Cameras Using Geometric Solvers},
  author    = {Zhao, Ji and Guan, Banglei and Liu, Zibin and Kneip, Laurent},
  booktitle = {IEEE/CVF Conference on Computer Vision and Pattern Recognition (CVPR)},
  year      = {2025}
}

@InProceedings{Messikommer2023,
  title     = {{Data-Driven Feature Tracking for Event Cameras}},
  author    = {Messikommer, Nico and Fang, Carter and Gehrig, Mathias and Scaramuzza, Davide},
  booktitle = {IEEE/CVF Conference on Computer Vision and Pattern Recognition (CVPR)},
  year      = {2023}
}

@InProceedings{Bonazzi2024,
  title     = {{Retina: Low-Power Eye Tracking with Event Camera and Spiking Hardware}},
  author    = {Bonazzi, Pietro and Bian, Sizhen and Lippolis, Giovanni and Li, Yawei and Sheik, Sadique and Magno, Michele},
  booktitle = {IEEE/CVF Conference on Computer Vision and Pattern Recognition Workshops (CVPRW)},
  year      = {2024}
}

@InProceedings{Bonazzi2025,
  author    = {Bonazzi, Pietro and Vogt, Christian and Jost, Michael and Khacef, Lyes and Paredes-Valles, Federico and Magno, Michele},
  title     = {{Towards Low-Latency Event-based Obstacle Avoidance on a FPGA-Drone}},
  booktitle = {IEEE/CVF Conference on Computer Vision and Pattern Recognition Workshops (CVPRW)},
  year      = {2025},
}

@InProceedings{Rebecq2018,
  title     = {{EMVS: Event-based Multi-View Stereo---3D Reconstruction with an Event Camera in Real-Time}},
  author    = {Rebecq, Henri and Gallego, Guillermo and Mueggler, Elias and Scaramuzza, Davide},
  booktitle = {European Conference on Computer Vision (ECCV)},
  year      = {2018}
}

@InProceedings{Vidal2018,
  title     = {{Ultimate SLAM? Combining Events, Images, and IMU for Robust Visual SLAM in HDR and High-Speed Scenarios}},
  author    = {Vidal, Antoni Rosinol and Rebecq, Henri and Horstschaefer, Timo and Scaramuzza, Davide},
  booktitle = {IEEE Robotics and Automation Letters (RA-L) with presentation at ICRA},
  year      = {2018}
}

@InProceedings{Zhou2018,
  title     = {{Semi-Dense 3D Reconstruction with a Stereo Event Camera}},
  author    = {Zhou, Yi and Gallego, Guillermo and Rebecq, Henri and Kneip, Laurent and Li, Hongdong and Scaramuzza, Davide},
  booktitle = {European Conference on Computer Vision (ECCV)},
  year      = {2018}
}

@InProceedings{Li2025,
  title     = {{Active Event-based Stereo Vision}},
  author    = {Li, Jianing and Zhang, Yunjian and Han, Haiqian and Ji, Xiangyang},
  booktitle = {IEEE/CVF Conference on Computer Vision and Pattern Recognition (CVPR)},
  year      = {2025}
}

@Article{Sopek2026,
  author    = {Sopek, Tamara and Zander, Fabian and Birch, Byrenn and Buttsworth, David},
  journal   = {Aerospace Science and Technology},
  title     = {{Development of neuromorphic event-based imaging spectroscopy for hypersonic flight observation}},
  year      = {2026},
  pages     = {111160},
  volume    = {168},
}

@InProceedings{Arja2023,
  title     = {{Density Invariant Contrast Maximization for Neuromorphic Earth Observations}},
  author    = {Arja, Sami and Marcireau, Alexandre and Balthazor, Richard L and McHarg, Matthew G and Afshar, Saeed and Cohen, Gregory},
  booktitle = {IEEE/CVF Conference on Computer Vision and Pattern Recognition Workshops (CVPRW)},
  year      = {2023}
}

@Article{Li2024,
  author    = {Li, Xiaopeng and Cheng, Shuaibo and Zeng, Zhaoyuan and Zhao, Chen and Fan, Cien},
  journal   = {Remote Sensing},
  title     = {{ERS-HDRI: Event-Based Remote Sensing HDR Imaging}},
  year      = {2024},
  number    = {3},
  pages     = {437},
  volume    = {16},
}

@Misc{Wildenstein2025,
  title  = {{STP-H12-VANTAGE: Combining Visual and Event-Based Sensing for Earth Observation}},
  author = {Wildenstein, Diego and Cannizzaro, Michael J and Peitzsch, Ian and Silbernagel, Linus and Poravanthattil, Joshua and Drum, Peter and Hofmeister, Mark and Bowman, Cole and Herzog, Natan and Gretok, Evan W and others},
  year   = {2025}
}

@Article{Eshraghian2023,
  author    = {Eshraghian, Jason K. and Ward, Max and Neftci, Emre O. and Wang, Xinxin and Lenz, Gregor and Dwivedi, Girish and Bennamoun, Mohammed and Jeong, Doo Seok and Lu, Wei D.},
  journal   = {Proceedings of the IEEE},
  title     = {{Training Spiking Neural Networks Using Lessons From Deep Learning}},
  year      = {2023},
  number    = {9},
  pages     = {1016--1054},
  volume    = {111},
}

@InProceedings{Xiao2025,
  title     = {{Spiking Meets Attention: Efficient Remote Sensing Image Super-Resolution with Attention Spiking Neural Networks}},
  author    = {Xiao, Yi and Yuan, Qiangqiang and Jiang, Kui and Huang, Wenke and Zhang, Qiang and Zheng, Tingting and Lin, Chia-Wen and Zhang, Liangpei},
  booktitle = {Conference on Neural Information Processing Systems (NeurIPS)},
  year      = {2025}
}

@InProceedings{Kucik2021,
  author    = {Kucik, Andrzej S. and Meoni, Gabriele},
  booktitle = {IEEE/CVF Conference on Computer Vision and Pattern Recognition Workshops (CVPRW)},
  title     = {{Investigating Spiking Neural Networks for Energy-Efficient On-Board AI Applications. A Case Study in Land Cover and Land Use Classification}},
  year      = {2021},
}

@InBook{Lemaire2020,
  author    = {Lemaire, Edgar and Millet, Philippe and Miramond, Benoît and Bilavarn, Sébastien and Saoud, Hadi and Naik, Alvin Sashala},
  pages     = {199--218},
  publisher = {Springer International Publishing},
  title     = {{Space Use-Case: Onboard Satellite Image Classification}},
  year      = {2020},
  booktitle = {Towards Ubiquitous Low-power Image Processing Platforms},
}

@Article{Smith2025,
  author    = {Smith, Stephen and Purcell, Cormac and Kuncic, Zdenka},
  journal   = {Scientific Reports},
  title     = {{Training-free AI for earth observation change detection using physics aware neuromorphic networks}},
  year      = {2025},
  number    = {1},
  volume    = {15},
}

@InProceedings{Kadway2023,
  author    = {Kadway, Chetan and Dey, Sounak and Mukherjee, Arijit and Pal, Arpan and Bézard, Gilles},
  booktitle = {2023 International Joint Conference on Neural Networks (IJCNN)},
  title     = {{Low Power \& Low Latency Cloud Cover Detection in Small Satellites Using On-board Neuromorphic Processors}},
  year      = {2023},
}

@Article{Bagchi2026,
  author    = {Bagchi, Samya and Anastasiou, Peter and Tetlow, Matthew and Chin, Tat-Jun and Latif, Yasir},
  journal   = {IEEE Transactions on Aerospace and Electronic Systems},
  title     = {{Event-based Star Tracking under Spacecraft Jitter: the e-STURT Dataset}},
  year      = {2026},
  pages     = {1--18},
}

@InProceedings{Dolan2026,
  title     = {{Sun-E: Dataset and Benchmark for Event-Based Sun Sensing}},
  author    = {Dolan, Sydney and Golkar, Alessandro},
  booktitle = {IEEE/CVF Winter Conference on Applications of Computer Vision (WACV)},
  year      = {2026}
}

@Article{Salah2024,
  author    = {Salah, Mohammed and Chehadah, Mohammed and Humais, Muhammad and Wahbah, Mohammed and Ayyad, Abdulla and Azzam, Rana and Seneviratne, Lakmal and Zweiri, Yahya},
  journal   = {IEEE Transactions on Instrumentation and Measurement},
  title     = {{A Neuromorphic Vision-Based Measurement for Robust Relative Localization in Future Space Exploration Missions}},
  year      = {2024},
  pages     = {1--12},
  volume    = {73},
}

@Article{Yao2025,
  author    = {Yao, Wei and Shen, Xin and Zhang, Guo and Lu, Zezhong and Wang, Jiaying and Song, Yanjie and Li, Zhiwei},
  journal   = {Swarm and Evolutionary Computation},
  title     = {{A spiking neural network based proximal policy optimization method for multi-point imaging mission scheduling of earth observation satellite}},
  year      = {2025},
  pages     = {101867},
  volume    = {94},
}

@InProceedings{Kahali2023,
  author    = {Kahali, Sayan and Dey, Sounak and Kadway, Chetan and Mukherjee, Arijit and Pal, Arpan and Suri, Manan},
  booktitle = {International Joint Conference on Neural Networks (IJCNN)},
  title     = {{Low-Power Lossless Image Compression on Small Satellite Edge using Spiking Neural Network}},
  year      = {2023},
}

@Article{Lunghi2025,
  author    = {Lunghi, Paolo and Silvestrini, Stefano and Dold, Dominik and Meoni, Gabriele and Hadjiivanov, Alexander and Izzo, Dario},
  journal   = {Astrodynamics},
  title     = {{Energy efficiency analysis of Spiking Neural Networks for space applications}},
  year      = {2025},
  number    = {6},
  pages     = {909--932},
  volume    = {9},
}

@Article{ElMaachi2025,
  author    = {El Maachi, Soukaina and Saadane, Rachid and Chehri, Abdellah},
  journal   = {Procedia Computer Science},
  title     = {{Intelligent Modeling of Soil Moisture Variability Using Remote Sensing and Spiking Neural Networks}},
  year      = {2025},
  pages     = {1372--1380},
  volume    = {270},
}

@Article{Tom2026,
  author    = {Tom, Manu and David, Cédric H. and Marlis, Kevin M. and Zimdars, Paul A. and Bonassies, Quentin and Wade, Jeffrey and Cerbelaud, Arnaud and Bonnema, Matthew and Pavelsky, Tamlin and Huang, Thomas},
  journal   = {Environmental Modelling \&; Software},
  title     = {Event-based serverless environmental modeling on the cloud: A pedagogical guide and river case study},
  year      = {2026},
  pages     = {106753},
  volume    = {196},
}

@InProceedings{Nishiguchi2024,
  title     = {{Event-based Vision Sensor Physics-Based Digital Twin for Tuning SSA Use}},
  author    = {Nishiguchi, Masashi and Frueh, Carolin and McReynolds, Brian},
  booktitle = {Advanced Maui Optical and Space Surveillance Technologies Conference (AMOS)},
  year      = {2024}
}

@Article{Roy2019,
  author    = {Roy, Kaushik and Jaiswal, Akhilesh and Panda, Priyadarshini},
  journal   = {Nature},
  title     = {{Towards spike-based machine intelligence with neuromorphic computing}},
  year      = {2019},
  number    = {7784},
  pages     = {607--617},
  volume    = {575},
}

@Article{Yang2026,
  author    = {Yang, Xu and Lei, Fuming and Tian, Na and Shi, Cong and Wang, Zhe and Yu, Shuangming and Dou, Runjiang and Feng, Peng and Qi, Nan and Wei, Zhongming and Liu, Jian and Wang, Kaiyou and Wu, Nanjian and Liu, Liyuan},
  journal   = {IEEE Journal of Solid-State Circuits},
  title     = {{A 10 000-Inference/s Bio-Inspired Spiking Vision Chip Based on an End-to-End SNN Embedding Image Signal Enhancement}},
  year      = {2026},
  number    = {3},
  pages     = {1164--1180},
  volume    = {61},
}

@Article{Capogrosso2026a,
  title   = {{TinyML Enhances CubeSat Mission Capabilities}},
  author  = {Capogrosso, Luigi and Magno, Michele},
  journal = {arXiv preprint arXiv:2603.20174},
  year    = {2026}
}

@Article{Capogrosso2024,
  author    = {Capogrosso, Luigi and Cunico, Federico and Cheng, Dong Seon and Fummi, Franco and Cristani, Marco},
  journal   = {IEEE Access},
  title     = {{A Machine Learning-Oriented Survey on Tiny Machine Learning}},
  year      = {2024},
  pages     = {23406--23426},
  volume    = {12},
}

@Misc{Esa2024,
  author       = {{The European Space Agency (ESA)}},
  title        = {{Neuromorphic AI Onboard}},
  year         = {2024},
  note         = {Accessed: 2026-03-24}
}

@Article{Cuadrado2023,
  author    = {Cuadrado, Javier and Rançon, Ulysse and Cottereau, Benoit R. and Barranco, Francisco and Masquelier, Timothée},
  journal   = {Frontiers in Neuroscience},
  title     = {Optical flow estimation from event-based cameras and spiking neural networks},
  year      = {2023},
  volume    = {17},
}

@InProceedings{Shi2024,
  title     = {{SpikingResformer: Bridging ResNet and Vision Transformer in Spiking Neural Networks}},
  author    = {Shi, Xinyu and Hao, Zecheng and Yu, Zhaofei},
  booktitle = {IEEE/CVF Conference on Computer Vision and Pattern Recognition (CVPR)},
  year      = {2024}
}

@InProceedings{Gehrig2019,
  title     = {{End-to-End Learning of Representations for Asynchronous Event-Based Data}},
  author    = {Gehrig, Daniel and Loquercio, Antonio and Derpanis, Konstantinos G and Scaramuzza, Davide},
  booktitle = {IEEE/CVF International Conference on Computer Vision (ICCV)},
  year      = {2019}
}

@Article{Shi2025,
  author    = {Shi, Yangsi and Li, Miao and Chen, Nuo and Luo, Yihang and He, Shiman and An, Wei},
  journal   = {Remote Sensing},
  title     = {{Sparse-Gated RGB-Event Fusion for Small Object Detection in the Wild}},
  year      = {2025},
  number    = {17},
  pages     = {3112},
  volume    = {17},
}

\end{document}